\documentclass[10pt,twocolumn,letterpaper]{article}

\usepackage[pagenumbers]{cvpr} 

\usepackage{graphicx, amsmath, amssymb, caption, subcaption, multirow, overpic, textpos}
\usepackage[table]{xcolor}
\usepackage[british, english, american]{babel}
\usepackage[pagebackref=false, breaklinks=true, letterpaper=true, colorlinks,
            citecolor=citecolor, linkcolor=linkcolor, bookmarks=false]{hyperref}
\definecolor{citecolor}{HTML}{0071BC}
\definecolor{linkcolor}{HTML}{ED1C24}

\def\cvprPaperID{**}
\def\confName{****}
\def\confYear{****}

\newlength\savewidth

\renewcommand{\paragraph}[1]{\vspace{1.25mm}\noindent\textbf{#1}}

\newcolumntype{x}[1]{>{\centering\arraybackslash}p{#1pt}}
\newcolumntype{y}[1]{>{\raggedright\arraybackslash}p{#1pt}}
\newcolumntype{z}[1]{>{\raggedleft\arraybackslash}p{#1pt}}

\newcommand{\app}{\raise.17ex\hbox{$\scriptstyle\sim$}}

\definecolor{deemph}{gray}{0.6}

\definecolor{baselinecolor}{gray}{.9}

\newcommand{\authorskip}{\hspace{2.5mm}}

\begin{document}
\title{
\vspace{-1mm}\Large Research on Patch Attentive Neural Process\vspace{-3mm}}
\author{
 Xiaohan Yu$^{1}$ \authorskip Shaochen Mao$^{1}$    \\[2mm]
 College of Command and Control Systems, Army Engineering University of PLA\vspace{-4mm}
}
\maketitle

\begin{abstract}
Attentive Neural Process (ANP) improves the fitting ability of Neural Process (NP) and improves its prediction accuracy, but the higher time complexity of the model imposes a limitation on the length of the input sequence. Inspired by models such as Vision Transformer (ViT) and Masked Auto-Encoder (MAE), we propose Patch Attentive Neural Process (PANP) using image patches as input and improve the structure of deterministic paths based on ANP, which allows the model to extract image features more accurately and efficiently reconstruction.
\end{abstract}

\section{Introduction}
\label{sec:intro}

Neural processes (NPs) \cite{art1} are a class of models that take advantage of the computational advantages of deep neural networks to mimic the function of Gaussian process regression (GPR) \cite{art2}. NPs can learn the distribution over functions with $O(n)$ computational complexity and make flexible predictions based on contextual data. Compared with Gaussian process regression, NPs can be trained to obtain implicit kernel functions from data without artificially designing fixed kernel functions, and has now been widely used in fields such as data complementation \cite{art3}, video prediction \cite{art4}, and robot path planning \cite{art5}.

\begin{figure*}[h]\centering
\includegraphics[width=0.75\linewidth]{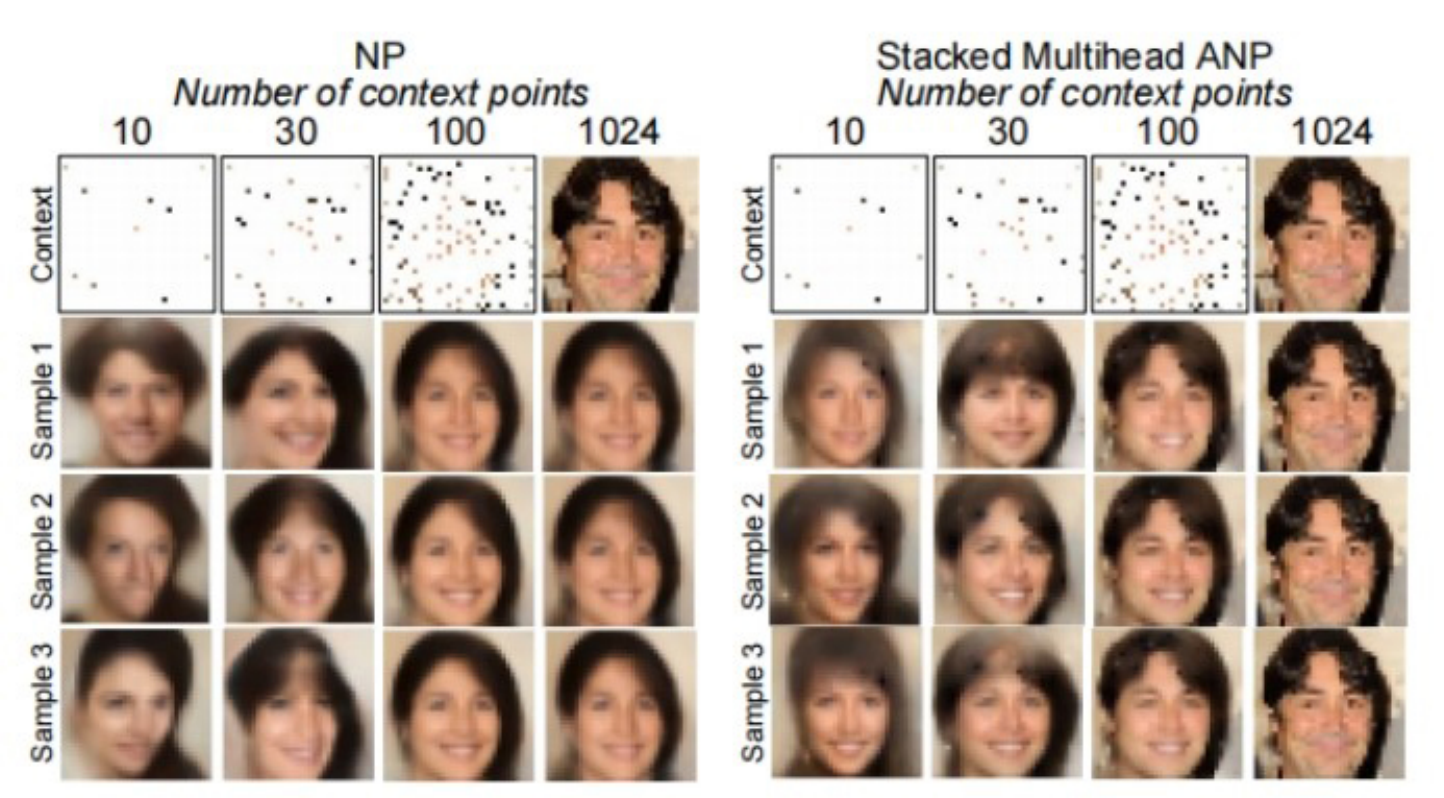}
\caption{Complementation results of (a)NP and (b)ANP.}
\label{fig:fig1}
\end{figure*}

Although the NPs do not focus on image processing, the model proponents applied NP to the image reconstruction task in a 2-D function regression experiment, as shown in Figure 1(a), where the NP achieved inference of the complete image only using less than 10\% of the pixels. At the same time, the researchers also found that it was difficult to reconstruct the original image even given all pixels as contextual data, i.e., the model is deficient in fitting the contextual data. Hyunjik Kim et al. suggested that fitting deficiency may be due to taking the mean value between context representations after aggregation, and introduced the attention mechanism \cite{art6, art7} into NP. They referred to Image Transformer and constructed Attentive Neural Process (ANP) \cite{art8} to solve this deficiency, and the reconstruction effect is shown in Figure 1(b).

However, the attention mechanism substantially improves the fitting and prediction ability of NP, its computational complexity also limits the length of the input sequence. In Figure 1, ANP only takes 32×32 image pixels as input, and the pixel sequence of each image is about 1000. However, when facing higher resolution images (e.g., 600×600, 800×800), it is difficult for ANP to effectively cope with the excessively long pixel sequences if it continues to use the image pixels as input elements. Facing the limitation of attention mechanism in image processing, Wang et al. designed the structure of the non-local neural network \cite{art9}, using the feature map of a convolutional neural network as the input of attention module, which is more convenient for processing compared with the original image; Ramachandram proposed isolated attention and Huiyu Wang et al. proposed axial attention \cite{art10} to solving the higher computational complexity, but the above two schemes use specialized attention, which is difficult to scale effectively on hardware.

Cordonnier et al. proposed to split the image into 2 × 2 patches \cite{art11}, with patches instead of pixels as input and validated on the Cifier dataset; Subsequently, Alexey et al. proposed the Vision Transformer (ViT) \cite{art12} to split the image into 16 × 16 patches with a linear mapping, which greatly reduces the length of the input sequence. And each patch is the inputs of the self-attentive encoder. Inspired by this, we redesigned the model structure of Attentive Neural Process (ANP) and called it the Patch Attentive Neural Process (PANP). The model uses the linear mapping of patches as input, which relieves ANP from processing high-dimensional inputs. At the same time, the encoder adopts a ViT-base structure, which facilitates the extraction of a small amount of abstract information of context data and improves the model feature extraction ability to predict the target data more accurately.

\section{Background}\label{sec:related}

The task we deal with in this paper is the regression based on meta-learning. The dataset is generated by Gaussian process with a fixed kernel, and consists of function family: $f:X\to Y$. Any $f_{}(x)$ in the function family is a sample function of the Gaussian process, that is, $f_{}(x)\sim \mathcal{GP}$. Considering the observed noise $\epsilon\sim N(0,\sigma^2)$, then $Y\tilde{\ }\mathcal{N}(f(X),{{\sigma }^{2}})$. And the task can be defined as predicting the target outputs ${{Y}_{T}}:={{\{{{y}_{t}}\}}_{t\in T}}$ corresponding to target inputs ${{X}_{T}}:={{\{{{x}_{t}}\}}_{t\in T}}$, given an arbitrary number of observed data $({{X}_{C}},{{Y}_{C}}):={{\{({{x}_{c}},{{y}_{c}})\}}_{c\in C}}$ (also called contexts).

To address this problem, Garnelo et al. propose Conditional Neural Process (CNP) \cite{art13} with deterministic path and Neural Process (NP) with latent path. Both of these two models mentioned above satisfy exchangeability and consistency, and other members of the NP family are defined using either just the deterministic path, just the latent path, or both.

\begin{figure*}[h]\centering
\includegraphics[width=0.75\linewidth]{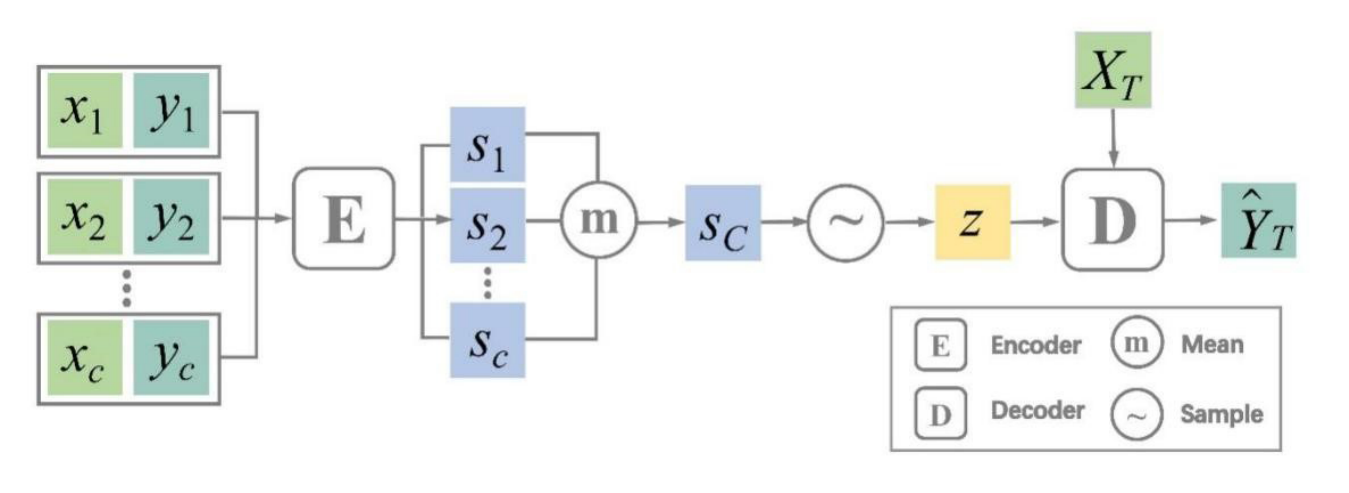}
\caption{Neural process model.}
\label{fig:fig2}
\end{figure*}

To represent a Gaussian process by NP, we assume that the Gaussian process $F(x,z)$ can be parameterized by a high-dimensional vector $z$ called the latent variable, and the stochasticity of the Gaussian process is derived from $z$. Sampling the latent variable $z$ yields a deterministic function $F(x,{{z}_{i}})={{f}_{i}}(x)$, and realizes global sampling of functions, where $f(x)$ is a fixed, learnable function and $z$ is a global latent variable that subjects to Gaussian distribution. The model structure is shown in Figure 2. And distribution of the targets can be represented as:
\begin{equation}
p({{Y}_{T}}|{{X}_{T}},{{X}_{C}},{{Y}_{C}}):=\int{p({{Y}_{T}}|{{X}_{T}},z)}q(z|{{s}_{C}})dz 
\end{equation}	
where the distribution $q(z|{{s}_{C}})=\mathcal{N}(z|\mu ({{s}_{C}}),diag{{[\sigma ({{s}_{C}})]}^{2}})$ of the latent variable $z$ is parametrized by $s_{C}$; The contexts $(X_{C},Y_{C})$ are encoded by the encoder $E$ and summed to the mean ${{s}_{C}}={}^{1}/{}_{\text{ }|\text{ C }|\text{ }}\begin{matrix}
   \sum\nolimits_{\forall c\in C}{{{s}_{c}}}  \\
\end{matrix}$, and the likelihood $p({{Y}_{T}}|{{X}_{T}},z)$ is modeled by a Gaussian distribution that can be factored as follows:
\begin{equation}
\begin{split}
p({{Y}_{T}}|{{X}_{T}},{{s}_{C}})\\
:&=\prod\limits_{t\in T}{\mathcal{N}({{y}_{t}}|\mu ({{x}_{t}},{{s}_{C}}),diag[\sigma {{({{x}_{t}},{{s}_{C}})}^{2}}])}
\end{split}
\end{equation}	
And in the Equation (2), the decoder $D$ predicts mean $\mu (\cdot ,\cdot )$ and variance $\sigma (\cdot ,\cdot )$by modeling the set of equations about $x_{t}$ and ${{s}_{C}}$.

Inspired by the Variational Autoencoders (VAE) \cite{art14}, we follow the ideas of related latent variable models that use the reparameterization trick to represent the latent variable $z$ and optimize the likelihood function by amortized variational inference, so that realize the parameters learning of the encoder and decoder. This gives the evidence lower-bound (ELBO): 
\begin{equation}
\begin{split}
logp(\left. {{Y}_{T}} \right|{{X}_{T}},{{X}_{C}},{{Y}_{C}})&\ge {{E}_{q(\left. z \right|{{s}_{T}})}}[\log p(\left. {{Y}_{T}} \right|{{X}_{T}},z)]\\
&-KL(\left. q(z\left| {{s}_{T}} \right.) \right\|q(z\left| {{s}_{C}} \right.))
\end{split}
\end{equation}
Since we consider that the contexts and the targets are generated in the same nonlinear function $h(x)$ in a Gaussian process, NP learns to predict the targets accurately by increasing the ${{E}_{q(\left. z \right|{{s}_{T}})}}[\log p(\left. {{Y}_{T}} \right|{{X}_{T}},{{r}_{C}},z)]$ term on the one hand, and on the other hand, the former term is regularized by the $KL(\left. q(z\left| {{s}_{T}} \right.) \right\|q(z\left| {{s}_{C}} \right.))$ term that controls the difference between the distributions of the contexts and the targets. And the $KL$ term ensures that the distributions of them are as close as possible. As such, the regression function $f(x)$ is close to the objective function $h(x)$, and we can predict output ${{\hat{y}}_{t}}=f({{x}_{t}})+\epsilon $ corresponding to the target input $x_{t}$ accurately.
In practice, although NP exhibits attractive attributes such as linear complexity, making predictions conditioned on arbitrarily sized contexts, and permutation invariance, the dimensionality of the latent variable still limits the expressiveness of the model. Additionally, the same weight of each context point makes it difficult for the decoder to distinguish which context points provide relevant information for a given target prediction. These defects above lead to underfitting and unsatisfactory target prediction ability of NP.
To improve these shortcomings, Hyunjik Kim et al. draw inspiration from the kernel of the Gaussian process to determine the similarity of two points, introduce the attention mechanism to NPs, and propose Attentive Neural Process (ANP). ANP uses both deterministic path and latent path, and significantly improves the accuracy of prediction by utilizing multi-headed attention to measure the relevance of the target representations to the context representations. One of the imperfections that detract from the positives of ANP is applying cross-attention to compute weights for all contexts, which leads to $O(n^2)$ computational complexity, and affects the application of the model to high-dimensional input tasks.

\vspace{1mm}\section{Patch Attentive Neural Process}\vspace{0.5mm}
\label{sec:approach}

Neural Process (NP) is divided into two parts: the Encoder parameterizes the Gaussian distribution of the latent variable $z$ and extracts the data features, and the Decoder predicts the target output based on the latent variable $z$ and the target input $x_{t}$. The latent variable $z$ is of global property, i.e., the property of the variable corresponding to all points in any sample are invariant, so the predictive ability of the model and the ability to fit the sample contextual data have a strong dependence on the latent variable. However, the representation ability of the latent variable is limited, and experiments have demonstrated that infinitely increasing the dimensionality of the latent variable does not consistently improve the predictive ability of NP.

The Attentive Neural Process (ANP), on the other hand, uses a deterministic path and introduces a cross-attention mechanism in this path. In the 2-D regression task (image reconstruction), cross-attention uses the coordinate embeddings of the target pixels corresponding to the target pixels as Queries, the coordinate embeddings of the context pixels as Keys, and the context pixel embeddings as Values to generate the corresponding representational vectors $r*$ for each target pixel. Therefore, in the ANP, the decoder receives the representation vectors generated from the deterministic path in addition to the global latent variable and target coordinates, avoiding the structure that the target output depends only on the latent variable $z$. However, this pixel-by-pixel computation of representations also imposes a limitation on the length of the input sequence of samples.

\begin{figure*}[h]\centering
\includegraphics[width=0.75\linewidth]{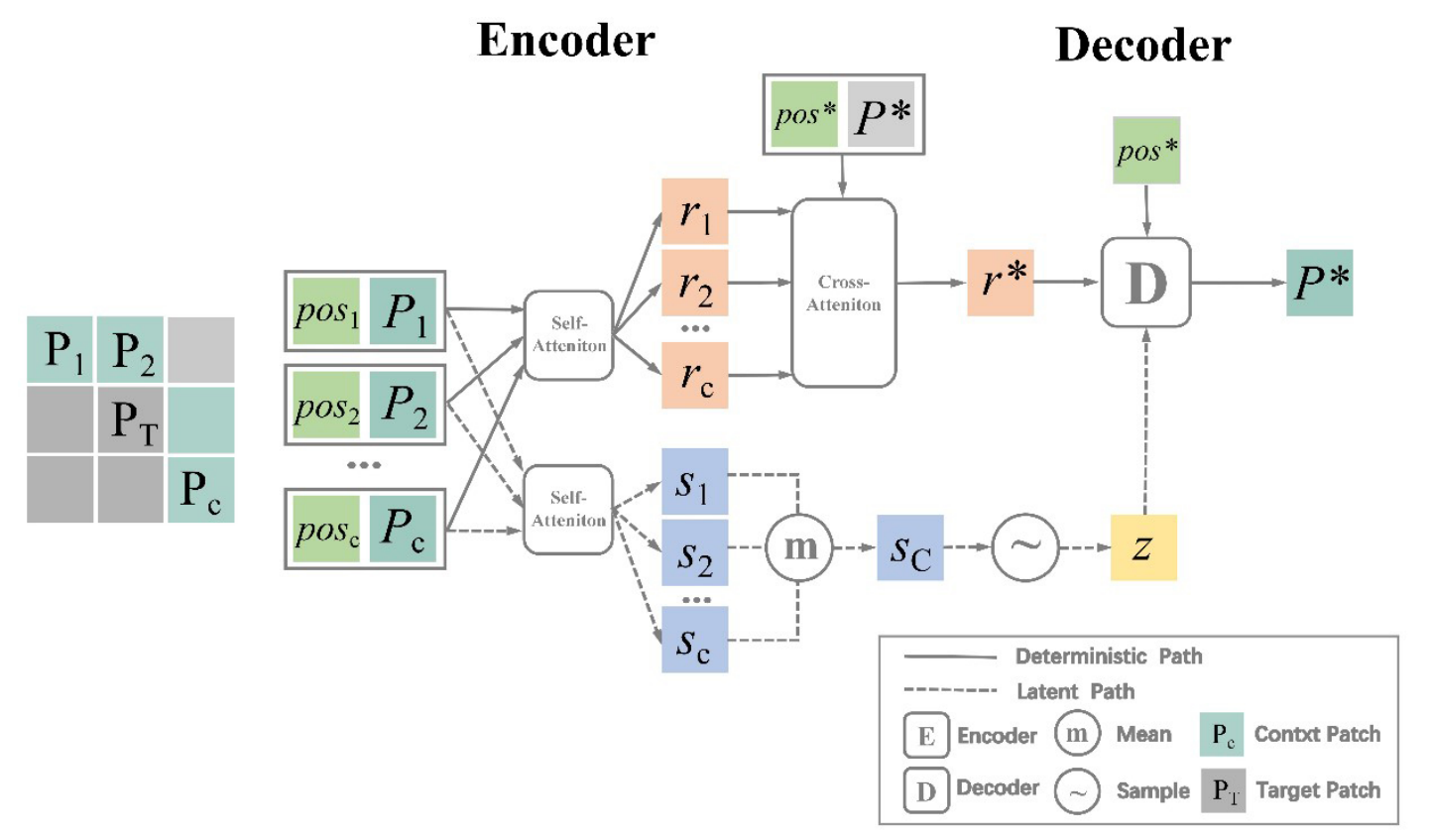}
\caption{Patch Attentive Neural Process model.}
\label{fig:fig3}
\end{figure*}

Input image in the field of computer vision has the same input structure as the Neural Process (NP) in 2D regression task, and had faced the same dilemma as NP applying the attention mechanism, i.e., the input sequence length is limited. Recent works, such as Vision Transformer, Masked Auto-Encoder \cite{art15}, etc., have brought new ideas for introducing attention mechanisms to neural processes and avoiding the input sequence length limitation problem. In this paper, we are inspired by this idea and improve the deterministic path of the attentive neural process to build a patch attentive neural process model to solve the defect that the 2D regression task of the neural process can only process small size images. The structure of PANP model is shown in Figure 3.

Specifically, we slice the image into equal-sized patches, and then convolve each patch and share the parameters of the convolution kernel to obtain the representation vector of each patch. Subsequently, the representation vectors corresponding to the context patches and the position embedding of the patches are used as inputs for feature extraction via deterministic and latent paths respectively. In the deterministic path, the stacked self-attention encoder will extract the context data features and output the corresponding representations $r_{i}$; all the context representations $r_{i}$ obtain the global vector $r_{C}$ by the mean operation, and $r_{C}$ as the representations of the unknown patch $P*$, i.e., $P*$ combined with the position embedding $pos*$ of the target patch as the query, and all the context representations $r_{i}$ are input as the key and value of the  cross-attention module, thus obtains the representation $r*$ of the target patch. In the latent path, the global representation $s_{C}$ is obtained by averaging the features $s_{i}$ extracted from the attention encoder, and then the global latent variable $z$ is obtained by sampling in the distribution of the latent variable. At the decoder side, the global latent variable $z$ is decoded with the corresponding representation $r*$ of the target patch and the position embedding $pos*$ as input to obtain the predicted $P*$. The decoder is structured as a multilayer perceptron and uses Gelu as the activation function.

The PANP overcomes the dilemma of too long pixel sequences by dividing the context data into equal-sized patches, while the stacked self-attention modules in the deterministic and latent paths facilitate the extraction of context features, and the deterministic representation $r*$ is fed into the decoder together with the global latent variable $z$, which helps to generate more accurate target prediction values.

{
\fontsize{8.2pt}{9.84pt}\selectfont
\bibliographystyle{unsrt}\bibliography{panp}}

\clearpage
\end{document}